\documentclass[runningheads]{llncs}
\usepackage{graphicx}
\usepackage{amsmath}
\usepackage[thinc]{esdiff}
\begin{document}
\title{StyPath: Style-Transfer Data Augmentation For Robust Histology Image Classification}
%
%
%

\author{Pietro Antonio Cicalese\inst{1,2} \and Aryan Mobiny\inst{1} \and Pengyu Yuan\inst{1} \and Jan Becker\inst{3} \and Chandra Mohan\inst{2} \and Hien Van Nguyen\inst{1}}
%
%
\authorrunning{P. A. Cicalese et al.}
%
\institute{Department of Electrical Engineering, University of Houston, Houston, TX, USA \\
\and Department of Biomedical Engineering, University of Houston, Houston, TX, USA \\ \email{pcicalese@uh.edu} \and Institute of Pathology, University Hospital of Cologne, Cologne, Germany}

\maketitle              

\begin{abstract}
The classification of Antibody Mediated Rejection (AMR) in kidney transplant remains challenging even for experienced nephropathologists; this is partly because histological tissue stain analysis is often characterized by low inter-observer agreement and poor reproducibility. One of the implicated causes for inter-observer disagreement is the variability of tissue stain quality between (and within) pathology labs, coupled with the gradual fading of archival sections. Variations in stain colors and intensities can make tissue evaluation difficult for pathologists, ultimately affecting their ability to describe relevant morphological features. Being able to accurately predict the AMR status based on kidney histology images is crucial for improving patient treatment and care. We propose a novel pipeline to build robust deep neural networks for AMR classification based on StyPath, a histological data augmentation technique that leverages a light weight style-transfer algorithm as a means to reduce sample-specific bias. Each image was generated in $1.84 \pm 0.03$ seconds using a single GTX TITAN V gpu and pytorch, making it faster than other popular histological data augmentation techniques. We evaluated our model using a Monte Carlo (MC) estimate of Bayesian performance and generate an epistemic measure of uncertainty to compare both the baseline and StyPath augmented models. We also generated Grad-CAM representations of the results which were assessed by an experienced nephropathologist; we used this qualitative analysis to elucidate on the assumptions being made by each model. Our results imply that our style-transfer augmentation technique improves histological classification performance (reducing error from 14.8\% to 11.5\%) and generalization ability.

\end{abstract}

\section{Introduction}
Low intra and inter observer reproducibility in the analysis of histologic features has been a persistent concern amongst pathologists across various tissue types and diseases. Although concordance is generally within an acceptable range for diagnosis, the clinical adoption of certain scoring parameters has been hindered by disagreement between observers and insufficient empirical evidence. This has a negative impact on the reproducibility of any given histological diagnosis and thus complicates the path to both consistent and effective treatment, and the design of multicentric pharmaceutical studies. The degree of disagreement observed has been the topic of several publications; varying levels of experience, the number of scoring parameters, cutoff values, and the type of stains used to assess a biopsy have all been associated with variable diagnosis \cite{dasari2019systematic,koelzer2015tumor,martin2018interobserver,wilhelmus2015interobserver}. Discrepancies in staining quality within and between pathology labs have also been implicated as a cause for variability, with variations in stain intensity being associated with poor diagnostic performance \cite{adam2014banff}. The effect of this apparent stain bias is similar to the texture and context bias we observe when training Convolutional Neural Networks (or CNNs), which negatively impacts the generalization ability of the model.

Geirhos \textit{et al.} showed that CNNs trained on the ImageNet dataset are heavily biased towards texture and fail to match human performance in the silhouette and edge classification tasks \cite{geirhos2018imagenet}. These models can therefore appear to yield exceptional performance in real world applications, when they are actually depending on a single cue for classification. This poses a significant challenge in histological diagnosis tasks where each biopsy has unique texture and color characteristics, which may have a negative impact on performance and the model's ability to generalize. This issue is also exacerbated by the relatively small sample sizes of histology datasets; whereas large datasets are less likely to yield models with sample specific bias, a small dataset is likely to yield models with poor generalization ability. We believe that by combining texture and color information from various training samples, we could reduce this texture bias while increasing the classifier shape bias, ultimately yielding a more robust model. Our contributions are as follows:

\begin{itemize}
   \item We propose a sample and condition agnostic style-transfer data augmentation technique to improve histological training performance with a small glomerular-level Antibody Mediated Rejection (AMR) dataset. Our technique depends on both the moderated transfer of spatial relationships and style, yielding a more diverse training population that retains the desired target concepts.
   \item We use Bayesian inference to compensate for the ignorance of the model (due to the small dataset size) and to estimate a more reliable measure of performance. Our technique improves upon the prediction accuracy of the baseline model at a lower computational cost relative to similar techniques.
   \item Qualitative analysis and visual inspection by an experienced nephropathologist are provided to demonstrate that the StyPath augmented models are more robust, capturing information that would have otherwise been missed by the baseline model.
\end{itemize}


\subsubsection{Related Works}
The issue of histological stain normalization has been addressed by various research groups in both the medical and computer science communities. Adam \textit{et al.} proposed a single feature schema in the qualitative scoring of polyomavirus BK (BKV) nephropathy immunohistochemistry (IHC) slides and stain protocol standardisation practices as a means to reduce the inter-observer variance associated with differing stain intensities \cite{adam2014banff}. This solution is limited by the difficulty associated with novel scoring parameter adoption and the cost incurred by pathology labs to standardize their equipment and procedures. These kinds of limitations have ultimately led various groups to develop \textit{in silico} stain normalization techniques that could be used to improve upon quantitative histological image analysis. In the hemotoxylin and eosin (H\&E) data normalization task, Macenko \textit{et al.} proposed a deconvolution technique which effectively separates the hemotoxylin and eosin stains to then generate their respective normalized components \cite{macenko2009method}. Bejnordi \textit{et al.} developed a whole-slide image color standardizer that combines color and spatial information to categorize each pixel in a slide scan as a stain component \cite{bejnordi2015stain}. Bug \textit{et al.} proposed a style-transfer normalization algorithm which transfers pathology lab profiles between samples without altering spatial relationships \cite{bug2017context}. Interestingly, Tellez \textit{et al.} evaluated all three of these stain normalization techniques and showed that they had a largely negligible (or even negative) effect on CNN classification performance while adding a significant computational cost, prompting them to favor other stain augmentation techniques \cite{tellez2019quantifying}. It is important to note that the style-transfer algorithm they evaluated only transfers the color profiles between pathology labs, remaining largely agnostic to the sources of variability present within each lab and stained section. Shaban \textit{et al.} and BenTaieb \textit{et al.} also proposed conditional stain transfer techniques that utilized adversarial networks to generate new samples for each target domain \cite{bentaieb2017adversarial,shaban2019staingan}. While powerful, these techniques require explicitly defined domain labels; this prevents the techniques from capturing important variations that may be present within each predefined condition set. We were interested in developing a sample and condition agnostic style-transfer augmentation procedure that could be applied to any histological dataset to address all sources of stain variability. We also propose the moderated transfer of spatial relationships between samples; while prior techniques avoided altering the morphological appearance of each image, we hypothesized that slight image alterations through the style-transfer algorithm could be beneficial to the generalization capabilities of a model, especially when trained with a small dataset.


\section{Methodology}
\subsubsection{AMR dataset generation and annotation}
A total of 86 (38 non-AMR and 48 AMR, chronic active AMR, or chronic AMR) blood group ABO- compatible kidney transplant biopsies were randomly selected for processing and analysis, each fulfilling the minimum sample criteria ($\geq7$ glomeruli, $\geq1$ artery). Each paraffin embedded section was cut at 2$\mu{m}$ and stained with periodic acid-Schiff (PAS) in the same pathology lab over the span of two years. For each section, micrographs were taken from all non-globally sclerosed glomeruli that were at least four levels apart at a resolution of $1024\times768$ pixels. All segmented glomerular images were then annotated by an experienced nephropathologist using the Labelbox platform; label choices were given as either AMR, non-AMR, or inconclusive, yielding $1503$ conclusively labeled glomerular images ($1001$ non-AMR and $502$ AMR) \cite{labelbox}.

\subsubsection{Style-transfer} 
One can think of the style-transfer algorithm as a means to generate artificial images that combine the high-level semantic representation of one image with the low-level perceptual representation of another image. The semantic image (i.e. the content image) represents the objects that will be depicted in the generated image, while the perceptual image (i.e. the style image) characterizes simpler information (such as color and texture). The ability to generate these artificial images depends on the extraction and manipulation of feature maps from the filters of the selected layers of a CNN \cite{gatys2015neural}. An input image $\textbf{x}$ is effectively encoded in each layer of a CNN by the filter responses to image $\textbf{x}$, which allows us to extract various representations of the image depending on the layer. Feature maps from deeper layers of a CNN characterize complex concepts (such as cells and glomeruli) while those from shallow layers characterize simple concepts (such as edges and color). We say that a layer $l$ with $N_l$ unique filters therefore has $N_l$ unique feature maps of dimensions $M^{hw}_{l}$, corresponding to the height multiplied by the width of each feature map. We can then say that the filter responses within a given layer $l$ can be stored in a matrix $F_l \in R^{N_l \times M^{hw}_l}$ where $F_{l}^{ij}$ corresponds to the activation of the $i^{th}$ filter at position $j$ in layer $l$.

Suppose that we have a given content image $\textbf{x}_{cont}$ and an output image $\textbf{x}_{out}$; we define their respective feature responses in a given layer $l$ as $F^{cont}_l$ and $F^{out}_l$, respectively. We can then define the squared-error content loss $\lambda^{cont}$ for a given layer $l$ following

\begin{equation}
\lambda^{cont}_{l}(\textbf{x}_{cont},\textbf{x}_{out}) = \frac{1}{2}\sum_{ij}( F^{ij,out}_{l} - F^{ij,cont}_{l})^{2}.
\end{equation}

\noindent
We can thus define the derivative of the squared-error loss between each set of feature representations for a given layer $l$ following

\begin{equation}
\diff{\lambda^{cont}_{l}}{F^{ij,out}_{l}} =
 \begin{cases} 
      (F^{ij,out}_{l} - F^{ij,cont}_{l}), & F^{ij,out}_{l} > 0 \\
      0, & F^{ij,out}_{l} < 0 
   \end{cases},
\end{equation}

\noindent
which can then be used to compute the content gradient with respect to image $\textbf{x}_{out}$. We then generate a Gram matrix which takes the correlation between various filter responses as a means to capture the style of the desired style image $\textbf{x}_{sty}$. The generated Gram matrix is given by $G_l \in R^{N_l \times M_l^{h} \times M_l^{w}}$, where $G^{ij}_{l}$ represents the inner product between the vectorised feature map $i$ and $j$ in a given layer $l$ at position $k$. We define $G^{ij}_{l}$ following

\begin{equation}
G^{ij}_{l} = \sum_{k} F_{l}^{ik}F_{l}^{jk},
\end{equation}

\noindent
which we then use to match the style between $\textbf{x}_{out}$ and $\textbf{x}_{sty}$ by minimising the mean-squared distance between their respective Gram matrices. Let $G^{ij,out}_{l}$ and $G^{ij,sty}_{l}$ represent the Gram matrices from a given layer $l$ of $\textbf{x}_{out}$ and $\textbf{x}_{sty}$, respectively. We then say that the contribution of layer $l$ to the total style loss is given by

\begin{equation}
\lambda^{sty}_{l} = \frac{1}{4(N_{l})^{2}(M^{hw}_{l})^{2}} \sum_{ij} (G^{ij,out}_{l} -G^{ij,sty}_{l})^{2},
\end{equation}

\noindent
while total style loss is given by

\begin{equation}
\lambda^{sty}(\textbf{x}_{sty},\textbf{x}_{out}) = \sum_{l=0}^{L} w_{l}\lambda^{sty}_{l},
\end{equation}

\noindent
with $w_{l}$ corresponding to weighting factors of the contribution of each layer to the total loss (which we simply set equal to one divided by the number of active layers). We can then compute the derivative of $\lambda^{sty}_{l}$ following

\begin{equation}
\diff{\lambda^{sty}_{l}}{F^{ij}_{l}} =
 \begin{cases} 
       \frac{1}{(N_{l})^{2}(M^{hw}_{l})^2}(F^{ij}_{l})^{T}(G^{ij,out}_{l} -G^{ij,sty}_{l}) & F^{ij}_{l} > 0 \\
      0 & F^{ij}_{l} < 0 
   \end{cases},
\end{equation}

\noindent
thus allowing us to compute the style gradient with respect to $\textbf{x}_{out}$. Finally, to generate the style-transfer samples, we simply minimize both $\lambda^{cont}$ and $\lambda^{sty}$ following

\begin{equation}
\lambda^{tot}(\textbf{x}_{cont},\textbf{x}_{sty},\textbf{x}_{out}) = \alpha\lambda^{cont}(\textbf{x}_{cont},\textbf{x}_{out}) + \lambda^{sty}(\textbf{x}_{sty},\textbf{x}_{out}),
\end{equation}

\noindent
where $\alpha$ is used to scale down the content loss, allowing us to control the stylization of the generated image.

\subsubsection{Approximate Bayesian Inference via MC-Dropout}
Training a standard neural network parameterized by its weights is equivalent to generating a maximum likelihood estimation (MLE) of the network parameters, which yields a single set of parameters \cite{goodfellow2016deep}. Such a model generates point estimates for each testing sample it classifies and ignores any model uncertainty that may be present in the proper weight values. In medical applications, this can eventually mislead the physician into believing that a model is confident about a prediction that may actually be a lucky guess \cite{gal2015bayesian}. Model uncertainty, also known as epistemic uncertainty, is most prevalent when a model is trained using a small sample set, where irrelevant information may be abused by the model to improve performance \cite{gal2015bayesian,mobiny2019dropconnect}. It would therefore be more informative to generate a model that provides a probabilistic estimate of its predictions, which can then be used to estimate the model's level of uncertainty. This can be accomplished by generating a Bayesian Neural Network (BNN) model; it is possible to generate a prior distribution over the network's parameters, outputting a probability distribution which can be used to estimate class posterior probabilities for each testing sample \cite{neal2012bayesian}. One can then integrate over the class posterior probabilities to produce a predictive posterior distribution over the class membership probabilities, and measure dispersion over the predictive posterior to generate uncertainty estimates. However, BNN models are computationally intractable, which has prompted various groups to develop methods to approximate Bayesian inference. To generate such a model, Gal \textit{et al.} proved that a feed-forward neural network with a given number of layers and non-linearities can be equivalent to approximate variational inference in the deep Gaussian Process model when dropout is applied to all units \cite{gal2016dropout}. We can therefore use Monte Carlo (MC) dropout at test time to yield a Bernoulli distribution over the weights of a CNN to generate an approximation of the posterior distribution without having to train additional parameters \cite{mobiny2019risk}. Through this technique, we can therefore quantitatively measure a more accurate estimate of each model's performance and describe their ability to generalize during the testing phase.

\subsubsection{Grad-CAM Visualization}
Being able to visualize and interpret the classification criteria being used by a CNN is critical to both confirming the assumptions being made by the classifier and learning from the classifiers decisions. Selvaraju \textit{et al.} showed that visual explanations for each model prediction could be generated by using gradient information flowing into the last convolutional layer of a CNN, capturing the semantic information being used to classify a given sample \cite{selvaraju2017grad}. These extracted Gradient-weighted Class Activation Maps (Grad-CAMs) could thus be used to depict a high-level representation of the classifiers decision making process. To further understand how both the baseline and StyPath augmented models drew their conclusions, we chose to generate Grad-CAM representations of each testing set prediction, which were then assessed by an experienced nephropathologist.

\section{Results and Discussion}

\begin{figure}[!t]
\centering
\includegraphics[width=\textwidth]{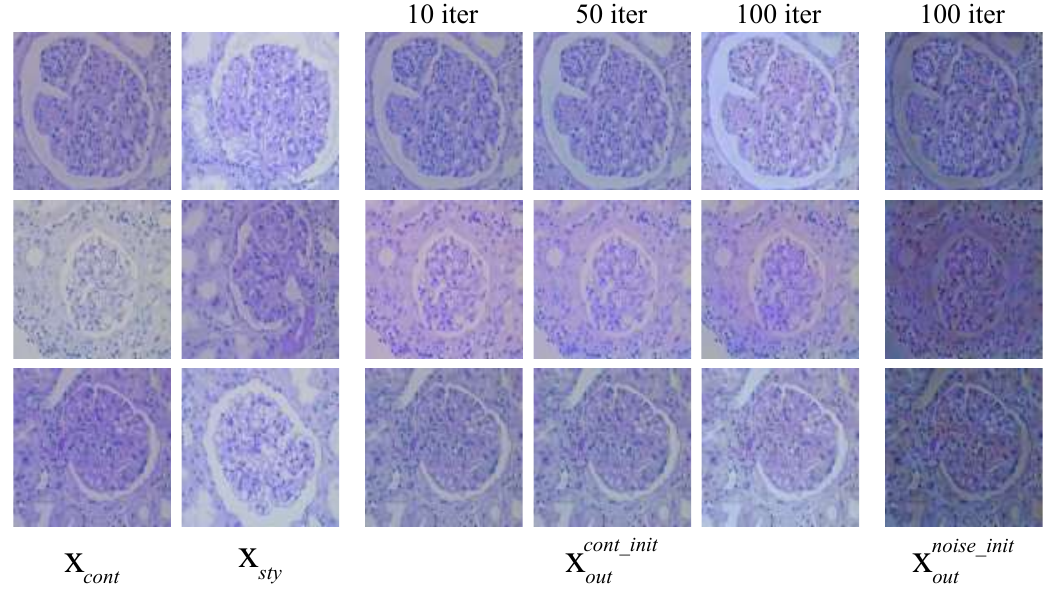}
\caption{Comparison of content and random initialization results. We note that output images initialized as the content image retain their morphological characteristics while capturing color and texture from the style image within 100 iterations. Output images initialized as noise appear distorted and discolored, failing to retain content fidelity.}
\label{initialization}
\end{figure}

\subsubsection{Style-Transfer Hyper Parameters}
We elected to use the original VGG19 network during the style-transfer sample generation phase due to its light-weight yet powerful contextual representation ability \cite{simonyan2014very}. Following Gatys \textit{et al.}, we take the output of the fourth convolutional layer for content and the outputs of convolutional layers one through five for style \cite{gatys2015neural}. We chose to initialize $\textbf{x}_{out}$ to be an exact copy of $\textbf{x}_{cont}$, allowing us to retain high content image fidelity while reducing the number of iterations needed to produce a meaningful output (100 iterations to generate each sample, as shown in Fig. \ref{initialization}). This allowed us to generate each image in $1.84 \pm 0.03$ seconds using a single GTX TITAN V gpu and pytorch, making it faster than other popular data augmentation techniques. This also helps soften the transfer of spatial relationships, allowing only slight structural modifications to occur that do not alter the desired target concept. We elected to use an $\alpha$ value of $2\times10^{-4}$ following visual inspection of the generated samples by an experienced nephropathologist; this ensured that the generated samples retained the morphological characteristics of their respective content image while capturing the texture and color characteristics of their respective style image. When generating each $\textbf{x}_{out}$, we selected one content and style image at random from the training set, irrespective of their associated label or section of origin. The generated style-transfer images were assigned the label of their corresponding content image and then appended to the training set before each augmented training experiment.

\subsubsection{Bayesian Classification Analysis}
\begin{figure}[!t]
\centering
\includegraphics[width=\textwidth]{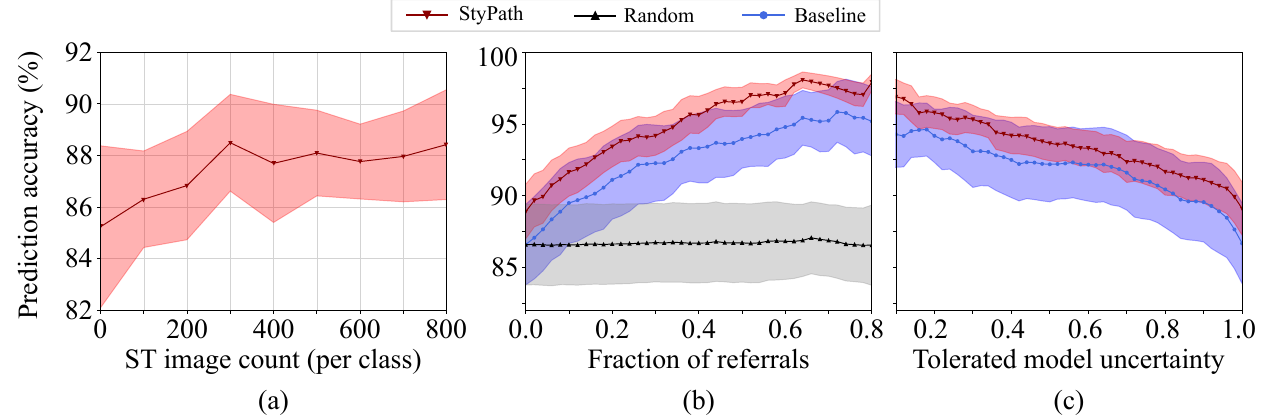}
\caption{Quantitative comparison of the baseline and StyPath augmented model performance. We note that the Bayesian performance of StyPath saturates after adding 300 Style-Transfer (ST) samples per class (a). When removing samples based on their uncertainty estimates, we note that the StyPath augmented model outperforms the baseline model while varying less across all five folds (b and c).}
\label{MC_Unc_filtering}
\end{figure}

During the classification task, we used the DenseNet-121 architecture pre-trained on the ImageNet dataset \cite{iandola2014densenet}. Both the original and StyPath augmented sample sets were resized to $256\times256$ pixels prior to being passed to each model; each classifier was trained for 200 epochs, using a batch size of 10, a drop rate of $0.1$ in all bottleneck blocks, and a learning rate of $10^{-4}$. Online augmentation was performed in all experiments for fair comparison between models; images had a $50\%$ chance of being flipped horizontally, being flipped vertically, having up to $30\%$ of their x and y axis cropped, and being rotated in either direction by up to $90$ degrees. Each testing fold in our five fold cross validation scheme consisted of original glomerular images; no images derived from the same section could be present in both the training and testing set folds. We gradually added an equal amount of randomly selected AMR and non-AMR style-transfer images to each training set to identify where performance saturated. We note that the baseline model achieves a weighted Bayesian classification accuracy of $85.2\%$ while the StyPath augmented model saturates around $88.5\%$ (after adding 300 style-transfer samples to each class, as shown in Fig. \ref{MC_Unc_filtering}a). To confirm that StyPath's performance increase had saturated, we also evaluated the model after adding $10000$ style-transfer samples to each class, which yielded a Bayesian classification accuracy of $88.2\%$.

\subsubsection{Epistemic Uncertainty Analysis}
We then filtered samples based on their normalized epistemic uncertainty values; we observe that the StyPath augmented model consistently outperforms the baseline model while its performance varies significantly less across all folds (as shown in Fig. \ref{MC_Unc_filtering}b and c). This result implies that StyPath augmented models have improved generalization ability, suggesting that the information being used by the model is more descriptive of the target concept.

\subsubsection{Qualitative Analysis using Grad-CAM}
\begin{figure}[!t]
\centering
\includegraphics[width=\textwidth]{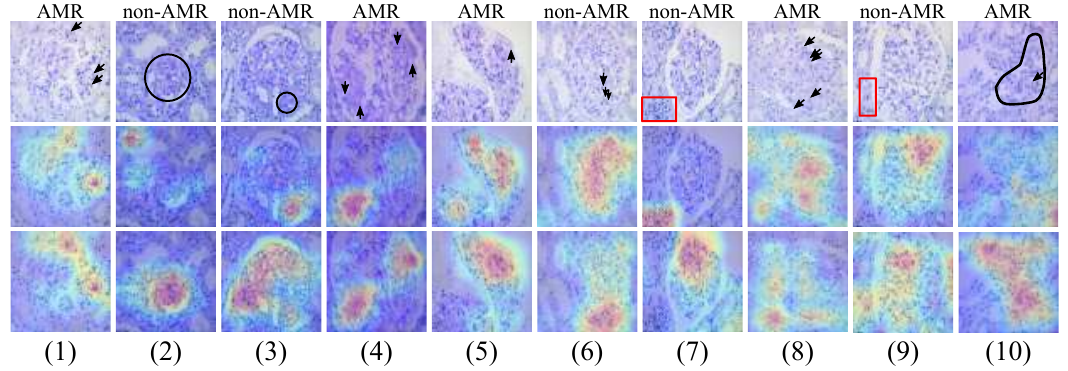}
\caption{Comparison of baseline (second row) and StyPath augmented (third row) Grad-CAM activations. Columns (1) through (5) represent success cases where the StyPath augmented prediction is correct and the baseline model prediction is incorrect, while columns (6) through (10) represent failure cases where the StyPath augmented prediction is incorrect and the baseline model prediction is correct. Black arrows in the original image (top row, labeled) are indicative of glomerulitis, mononuclear cells/infiltrates, and split glomerular basement membranes, while black and red outlines characterize regions of interest and misleading areas, respectively.}
\label{qualitative}
\end{figure}

We note that certain StyPath augmented model predictions caused the Grad-CAM activations to shift, widen, or narrow down to diagnostic features present in the glomeruli, correcting erroneous baseline model predictions (as shown in Fig. \ref{qualitative}, columns 1 through 5). While some failure cases indicate that the StyPath augmented model missed the desired concept (Fig. \ref{qualitative}, columns 8 and 9), the initial label (non-AMR) of Fig. \ref{qualitative}, column 6 was put into question upon observing the StyPath augmented Grad-CAM activation; the model detects an accumulation of intracapillary mononuclear cells that are indicative of glomerulitis (and thus AMR). We observe that the baseline model correctly classifies Fig. \ref{qualitative}, column 7 as non-AMR by counter intuitively focusing on an area of dense infiltrates in the periglomerular interstitium, which is not descriptive of non-AMR. Although the sample is misclassified, the StyPath augmented model instead focuses on the glomerular tuft, indicating that it detected the correct region of interest. We also note that generally, collapsed glomerular images seemed to pose a challenge for both the nephropathologist and the classifier, with both baseline and StyPath augmented models failing to focus on the areas indicative of AMR (Fig. \ref{qualitative}, column 10).

\section{Conclusions}
We present a novel histological data augmentation technique called StyPath, which generates new histological samples through the sample and condition agnostic transfer of both spatial relationships and style. We use Bayesian inference to evaluate the technique and show that it improves both the performance and generalization ability of the classifier at a low computational cost. We then generated Grad-CAM representations of both the baseline and StyPath augmented models for assessment by an experienced nephropathologist. This assessment showed that the augmented model tended to focus on morphologically relevant information which ultimately improved its classification accuracy. Our future works aim to compare the performance of StyPath to other SOTA techniques on larger multi-conditional datasets.

\vspace{5mm}
\noindent
\textbf{Acknowledgments.}
This research was supported by the National Science Foundation (NSF-IIS 1910973).

\bibliographystyle{splncs04}
\bibliography{references}

\pagebreak

\textbf{Supplementary Material}

\begin{figure}[!h]
\centering
\includegraphics[width=0.8\textwidth]{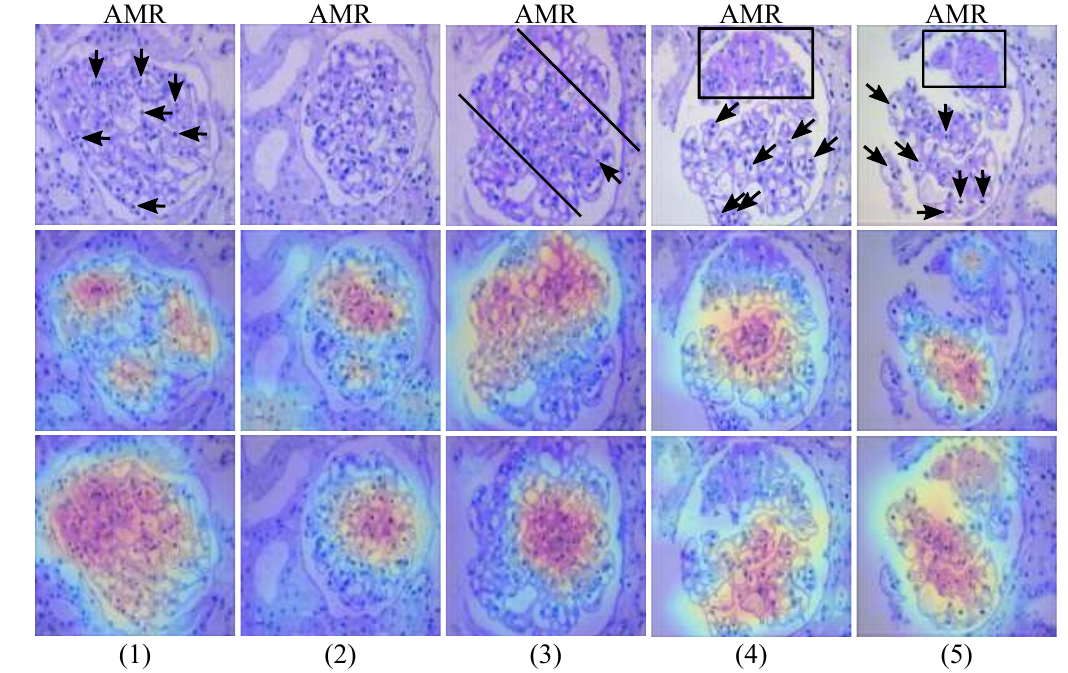}
\caption{Additional success cases. Black arrows in the original image (top row, labeled) are indicative of glomerulitis, mononuclear cells/infiltrates, and split glomerular basement membranes, while black outlines characterize regions of interest.}
\end{figure}

\begin{figure}[!h]
\centering
\includegraphics[width=0.8\textwidth]{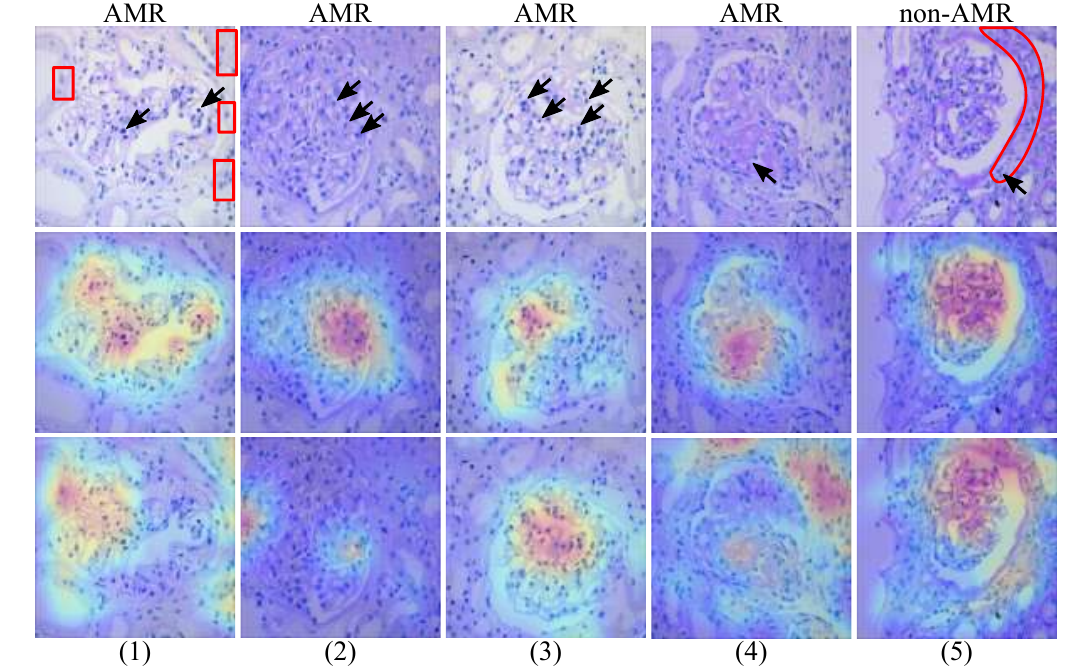}
\caption{Additional failure cases. Black arrows in the original image (top row, labeled) are indicative of glomerulitis, mononuclear cells/infiltrates, and split glomerular basement membranes, while black outlines characterize regions of interest.}
\end{figure}

%
%
%
%

\end{document}